\documentclass{article}

\usepackage{arxiv}

\usepackage[utf8x]{inputenc} 
\usepackage[T1]{fontenc}    
\usepackage{hyperref}       
\usepackage{url}            
\usepackage{booktabs}       
\usepackage{amsfonts}       
\usepackage{nicefrac}       
\usepackage{microtype}      
\usepackage{lipsum}
\usepackage{makeidx}         
\usepackage[bottom]{footmisc}

\usepackage{glossaries}
\usepackage[colorinlistoftodos,prependcaption,textsize=medium]{todonotes}
\usepackage[linesnumbered,ruled,lined,commentsnumbered]{algorithm2e}
\usepackage{multicol}
\usepackage{amsmath}
\usepackage{amssymb}
\usepackage{mathtools}
\usepackage{bm}
\usepackage{cleveref}
\usepackage{graphicx}
\usepackage{array}
\usepackage{multirow}
\usepackage{csquotes}
\usepackage{ctable}
\usepackage{subcaption}
\usepackage{float}

\usepackage{color, colortbl}
\emergencystretch=\maxdimen
\newtoggle{CAMERA}
\toggletrue{CAMERA}

\title{Avant-Satie! Using ERIK to encode task-relevant expressivity into the animation of autonomous social robots}

\author{
  Tiago Ribeiro\thanks{\protect\url{www.tiagoribeiro.pt}} \\
  INESC-ID \& \\Instituto Superior T\'{e}cnico\\
  University of Lisbon\\
  Portugal \\
  \texttt{me@tiagoribeiro.pt} \\
   \And
 Ana Paiva \\
  INESC-ID \& \\Instituto Superior T\'{e}cnico\\
  University of Lisbon\\
  Portugal \\
  \texttt{ana.paiva@inesc-id.pt} \\
}
\date{\vspace{-5ex}}
\begin{document}
\maketitle

\begin{figure}[htp]
	\centering
	\includegraphics[width=\columnwidth]{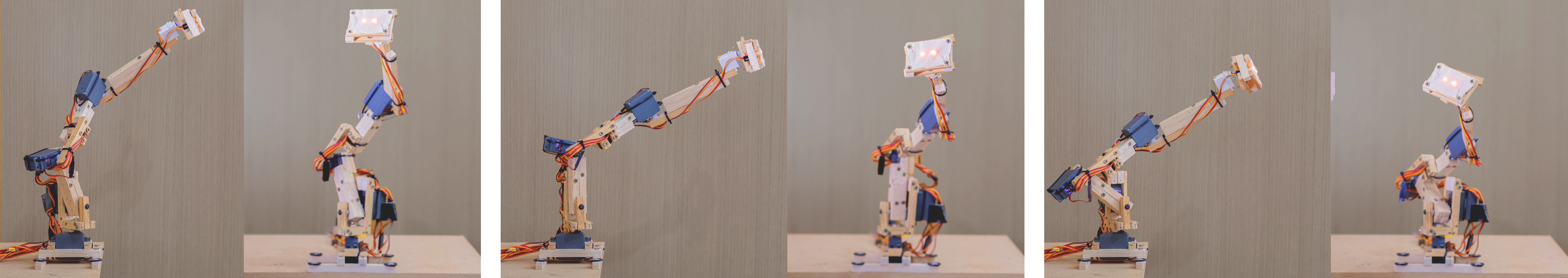}
	\caption{Pictures of the three different postures used by Adelino in the AvantSatie scenario (condition \textbf{C-ERIK}). 
		These are the actual output of ERIK as used in the game, while gazing forward, and not a set of static pre-designed postures. 
		Each pair shows a side view and its corresponding frontal view. The postures represented are: \textbf{left} - Neutral; \textbf{center} - Warm; \textbf{right} - Cold. }
	\label{fig:avantSatiePostures}
\end{figure}

\begin{abstract}
ERIK is an expressive inverse kinematics technique that has been previously presented and evaluated both algorithmically and in a limited user-interaction scenario.
It allows autonomous social robots to convey posture-based expressive information while gaze-tracking users.
We have developed a new scenario aimed at further validating some of the unsupported claims from the previous scenario.
Our experiment features a fully autonomous Adelino robot, and concludes that ERIK can be used to direct a user's choice of actions during execution of a given task, fully through its non-verbal expressive queues.


\end{abstract}


\newacronym{hri}{HRI}{Human-Robot Interaction}
\newacronym{cgi}{CGI}{Computer-Graphics}
\newacronym{ai}{AI}{artificial intelligence}
\newacronym{ik}{IK}{inverse kinematics}
\newacronym[longplural={Degrees of Freedom}]{dof}{DoF}{Degree of Freedom}

\section{Introduction}
Character animation is the process of breathing the illusion of life into visual drawings or renderings of the embodiment of a character that takes part in some kind of narrative \cite{thomas1995illusion,Goldberg2008}. 
The process however becomes much more difficult when applied to the field of autonomous social robots, given that the flow and especially the \textit{timeline} of the story is driven not only by the interaction between users and the \gls*{ai} \cite{Tomlinson2005}, but also because the spacial dimension of the interaction is also linked to the user's own physical motion and placement.

This challenge is remarkable enough that character animation for robots can be considered a new form of animation, which builds upon and extends the current concepts and practices of both traditional and \gls*{cgi} animation. Ribeiro \& Paiva have defined robot animation as \textit{the workflow and \textit{processes} that give a robot the ability of expressing identity, emotion and intention during autonomous interaction with human users} \cite{ribeiro2020phdthesis,ribeiro2019nutty}. 


One such common \textit{process} of robot animation that we are interested in is face-tracking, which directs a robot's gazing towards the face of the human with whom it is interacting.
For a simple robot, e.g., a neck with two \glspl*{dof}, face-tracking can be easily implemented by extracting a vertical and horizontal angle from the system's perception components (e.g. camera, Kinect), and using those to directly control individual motors of the robot's neck. 
Gazing behaviour can also be compound, featuring not only face-tracking, but also deictic queues towards surrounding objects, and in conjunction with other static or motive expression (e.g. posture of engagement, nodding in agreement), as is the case when using a manipulator-like robot who's end-effector takes on the expressive role of being the character's head (e.g. the Adelino robot, depicted in Figure \ref{fig:avantSatiePostures}).
In that case the compound-gazing process can also be used to express e.g. identity, emotion and intention, which are typically desirable for social robots \cite{Breazeal2008}.

In this paper we report how an existing inverse kinematics technique (ERIK) can be used in a real-world interactive application in which compound-gazing takes a major role, in order to allow an autonomous social robot to convey a recognizable intention-directed expression, and more importantly, to test its ability to direct the user's choice of action during execution of a given problem-solving task, fully through non-verbal expressive queues.

\section{Related Work}
\label{sec:related}
Robot animation builds on existing models and techniques from various fields. 
This section presents not all of them, but the ones we found most relevant for grounding work and inspiration.

Both Hoffman and Weinberg have created interactive robots that behave in musical and theatrical environments.
The AUR is a robotic desk lamp with 5 DoFs and an LED lamp which can illuminate in a range of the RGB color space \cite{Hoffman2008}.
It is mounted on a workbench and remotely controlled through a hybrid control system that allows it to be used for live puppeteering, in order to allow the robot to be expressive while also being responsive. 
Its motion was composed through several layers. 
The bottom-most one moves each DoF based on a pre-designed animation that was made specifically for the scene of the play. 
When set to establish eye contact, several specific DoFs would be overridden by an \gls*{ik} solution using CCD \cite{WangChen1991}. 
A final \textit{animacy} layer added smoothed sinusoidal noise, akin to breathing, to all DoFs, to provide a more lifelike motion to the robot.
Weinberg has also dedicated to the creation of robotic musical companions, such as Shimon, a gesture based musical improvisation robot created along with Hoffman \cite{HoffmanWeinberg2010}. 
Shimon plays a real marimba and its behaviour is a mix between his functionality as a musician, for which he plays the instrument in tune and rhythm, and being part of a band, for which he performs expressive behaviour by gazing towards his band mates during the performance.

Various expressive social robots have been created at MIT's MediaLab \cite{Gray2010}, in particular the AIDA, which is a friendly driving assistant for the cars of the future. 
AIDA interestingly delivers an expressive face on top of an articulated neck-like structure to allow to it move and be expressive on a car's dashboard. \cite{Aida}.


The use of animation principles has been developed and explored by various authors (e.g. \cite{Breemen2004a, Wistort2010, Mead2010, Gielniak2012}).
In particular, Takayama, Dooley and Ju \cite{Takayama2011} explored how the expression of intention leads to a sense of \textit{thought}, using the PR-2 robot \cite{PR2}, and featured a collaboration with a professional animator.
Later, Ribeiro \& Paiva have also proposed a list of Principles of Robot Animation, in both an early and revised version \cite{Ribeiro2012, ribeiro2019nutty}.
From all these authors we collect that \textit{Thought} and \textit{Intention} are two concepts that are central in character animation, and in the portrayal of the illusion of life.

The challenge of providing legible, predictable motion to autonomous collaborative robots was addressed by Dragan et al. \cite{Dragan2015}, who demonstrate the benefits of including such properties into motion planners.
Their technique however focuses on these properties in particular, and rely on motion-planning for trajectory generation.

\subsection{Expressive Inverse Kinematics}
Computing the motion of an articulated structure is commonly done through Forward Kinematics (FK) and Inverse Kinematics (IK).
Given a kinematic chain of $N$ segments $S_i$ connected through joints $J_i$, where each \textit{parent} joint allows its \textit{child} segment to rotate about an axis $R_i$, the process of Forward Kinematics is to calculate the resulting position and/or orientation of an Effector $S$ which is linked to any of the joints, given a set of angles applied to each joint; Inverse Kinematics is the calculation of the set of angles which, when applied to each joint, would bring the effector $S$ (as close as possible) to a target position and/or orientation $T$.
Due to space constraints we cannot fully describe the inner workings of all existing techniques, however those are already extensively described in both \cite{Buss2009} and \cite{Aristidou2018} which are excellent reads on the topic.
The most common techniques are based on iteratively calculating the inversion of the Jacobian matrix of the system, such as in the Jacobian transpose, DLS or SDLS techniques \cite{Buss2009, BussKim2005}.
Other popular techniques are the CCD \cite{WangChen1991} and FABRIK \cite{Aristidou2011a}, which take a geometrical approach to the problem instead of solving a matrix system.
These two methods in particular became popular in \gls*{cgi} as they produce good results with a low computational cost.
FABRIK in particular is highly flexible, extensive to various complex situations, and provides quick and naturally-looking results \cite{Aristidou2016}.

Previously, Ribeiro \& Paiva detailed and evaluated ERIK \cite{ribeiro2019erik}, an expressive kinematics technique that builds on both FABRIK and a variation of CCD (called \textit{BWCD}), which allows a robot to express a given posture towards a specified direction, given an arbitrary embodiment, in real-time and without requiring motion planning or prior training.
The technique is claimed to provide a flexible and extensible solution, untied to specific parameters such as an emotion or an interaction feature.
It was designed for real-time applications, with minimal authoring required, given that a single input posture can be expressed towards any direction without drastically compromising its underlying expressive intent, in a way that also allows it to convey the illusion of life.
However the technique's complex formulation also makes it difficult to assure its reliability and consistency, especially given that, besides its algorithmic evaluation performed in that paper, it was initially evaluated with users in \textit{Ahoy!}, a scenario in which participants remained static, and the robot was being controlled, to some extend, through a Wizard-of-Oz mechanism \cite{Ribeiro2017}.
\begin{figure}[t]
	\centering
	\includegraphics[width=0.6\columnwidth]{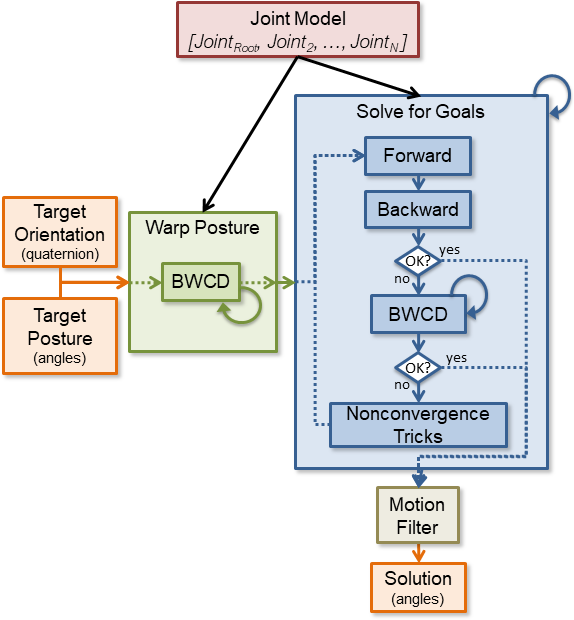}
	\caption{The ERIK Pipeline. The given Target Posture and Orientation are first warped using BWCD, so that the posture's end-point is aiming towards the Target Orientation, without enforcing joint limits. The result feeds into the Iterative portion, which, through various phases on each iteration, approximates the configuration to the target parameters following a geometric approach based on FABRIK, and returns the final solution (with limits enforced). The Joint Model contains the skeletal information and auxiliary operations. The final solution runs through a motion filter to ensure smooth, continuous output. \textit{(figure and text cited from \cite{ribeiro2019erik})}}
	\label{fig:erik_pipeline}
\end{figure}
The ERIK pipeline is shown in Figure \ref{fig:erik_pipeline} which also includes a brief description of its internals.
Its full description and algorithm is described in \cite{ribeiro2019erik}, and should be consulted by the interested reader.


\section{AvantSatie - The Piano Game Companion}
\label{sec:avantsatie}
After analysing the previous \textit{Ahoy!} scenario used to evaluate ERIK \cite{Ribeiro2017}, 
we became interested in studying its use in a more realistic and fully autonomous setting, by studying not only the robot's ability to convey a recognizable expression, but more importantly, its ability to direct the user's choice of action in a problem-solving task.
In the previous evaluation, the players were additionally standing still, which failed to validate the algorithm's ability to express a given posture towards different directions.
Furthermore, the queues were used to try to direct the players towards the correct answer in a pantomimic game, which took an open input (through speech), and required the use of \textit{Wizards} to listen and assess the player's answer, which then triggered the robot's reactive expression.
In that scenario, the robot seemed to be more like part of the riddle, than as part of a solution to it.
Our goal is to understand if the technique can actually be used in a collaborative scenario in which users solve a problem that is independent of the robot, but in which the robot's expressive behaviour may take a role in assisting the user, by providing expressive queues that facilitate the task.

The question we therefore posed to address is:
\begin{displayquote}
	\textit{Can ERIK provide Adelino with the ability to communicate non-verbal hints to a user about what action to take next during a particular task, while also conveying the illusion of life?}
\end{displayquote}

For that purpose we established the following requirements for a new evaluation scenario: 
the robot must be \textbf{fully autonomous}, in order to evaluate ERIK based on its actual response-time; 
users must be forced to \textbf{move around} in order to fully evaluate the expressive gaze-tracking behaviour; 
the task must be a non-subjective problem-solving one, and must be \textbf{solvable even without expressive queues} (hints) through a \textit{trial-end-error} method, however such hints should allow to solve it with significantly less \textit{errors}.

The previous scenario already featured Adelino (depicted in Figure \ref{fig:avantSatiePostures}), a highly expressive, articulated craft robot whose end-effector is used as a face, and was purposely built as a low-fidelity one with an \textit{organo-tech} look, in order to challenge the paradigms of robot design, and to seem more appealing to the arts'n'crafts and DIY\footnote{Do-It-Yourself \protect\url{http://en.wikipedia.org/wiki/Do_it_yourself}} communities.
One of our claims for building such a robot is that, although it exhibits some involuntary shakiness, as can be seen in the accompanying video, we consider that it can become part of its own unique \textit{character} and aesthetically even contribute to a sense of lifelikeness, as long as the animation system and techniques used (e.g. ERIK) are sturdy enough and can properly control its conveyed communicative intent beyond that issue.
This question of whether the communicative intent of the robot can be consistently conveyed beyond any shakiness or low-fidelity motion is therefore intrinsically part of our research question.

Based on these arguments, and because we thought Adelino's design was appropriate for our evaluation, we decided to use the same robot as in the previous scenario, given that our major concern was on the actual evaluation's activity design and conclusions, and not on the use of this particular robot.
In fact, if the claims do stand, then our opinion is that Adelino and \textit{AvantSatie} may also represent a breakthrough in robot and application design for HRI, and inspire future generations of robots.

\subsection{AvantSatie - Activity Design}
\begin{figure}[htbp]
	\centering
	\begin{subfigure}[t]{0.50\columnwidth}
		\includegraphics[width=1\columnwidth]{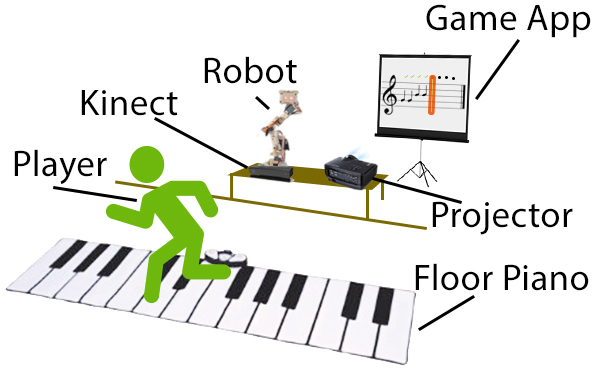}
		\caption{The set-up of AvantSatie.}~\label{fig:avantSatieSetup}
	\end{subfigure} ~
	\begin{subfigure}[t]{0.50\columnwidth}
		\includegraphics[width=1\columnwidth]{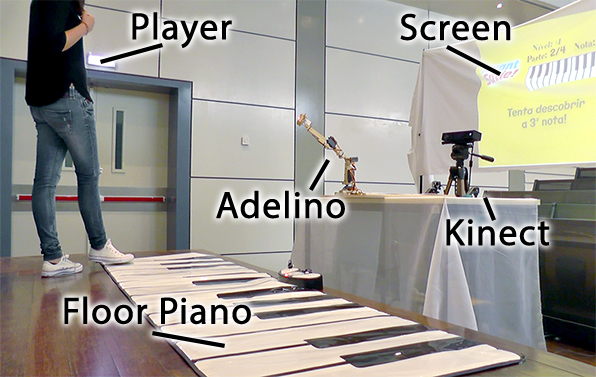}
		\caption{The shot of a study session.}~\label{fig:avantSatieStudy}
	\end{subfigure}
\end{figure}

AvantSatie is a pervasive game where players must discover the musical score of a piece using a floor piano where they can step on to play notes.
To help them in the game, they interact with an autonomous Adelino robot that will help them discover which notes compose the musical scores of two different pieces (each piece corresponds to a level).
By interacting with the piano, observing the robot and following the instructions, participants either adopt a trial-and-error process, or track the robot's hints, to discover each successive note.
The game's set-up is illustrated in Figure \ref{fig:avantSatieSetup} along with a shot of an actual experimental session (Fig. \ref{fig:avantSatieStudy}).

The fully autonomous robot enriches the setting of the game by providing the story as well as being socially present through compound-gazing.
Its gazing behaviour combines a face-tracking feature, and deictic gazing towards specific piano keys.
At the same time, it is expressive and can shape its posture while gazing, in order to convey hints to the player.
Through the understanding of such hints, players can play the game while minimizing the amount of mistakes performed through the trial-and-error nature of the gameplay. 
Yet, if they do not pay attention to the robot, or fail to understand it, their task in the game becomes much more difficult.
The game was designed and iteratively tested with users in pilot studies, in order to ensure that the instructions and gameplay were clear, instead of relying on initial instructions given by the experimenters, which could introduce biases.

As the game is about discovery, the scores and composition are initially unknown to the player.
They must therefore attempt to play keys on the piano until they find each correct note.
The robot's behaviour is fully non-verbal. Only a screen is projected behind it, providing basic instructions and progress (e.g. current level).
\begin{figure}[htp]
	\centering
	\includegraphics[width=0.55\columnwidth]{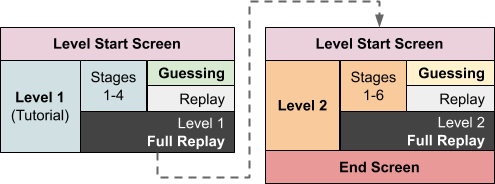}
	\caption{A diagram of the game-flow of Avant Satie. There are two levels, composed of a sequence of stages. Each stage is composed of a challenging activity (Guessing) followed by a rewarding activity (Replay), with each level ending with a greater reward (Full Replay).}~\label{fig:avantSatie_gameplay}
\end{figure}

The structure of the game-play is outlined in Figure \ref{fig:avantSatie_gameplay}.
There are two levels in the game, which correspond to two musical compositions, each one with distinct levels of difficulty. 
Each level (music) is split into stages, which we called Parts, as that term fits better in the context of the game (e.g. the music in Level~1 is composed of 4~Parts/Stages).
Each Part contains a sequence of one to six individual notes to be discovered one by one in the correct order.

\begin{figure}[htbp]
	\centering
	\includegraphics[width=1\columnwidth]{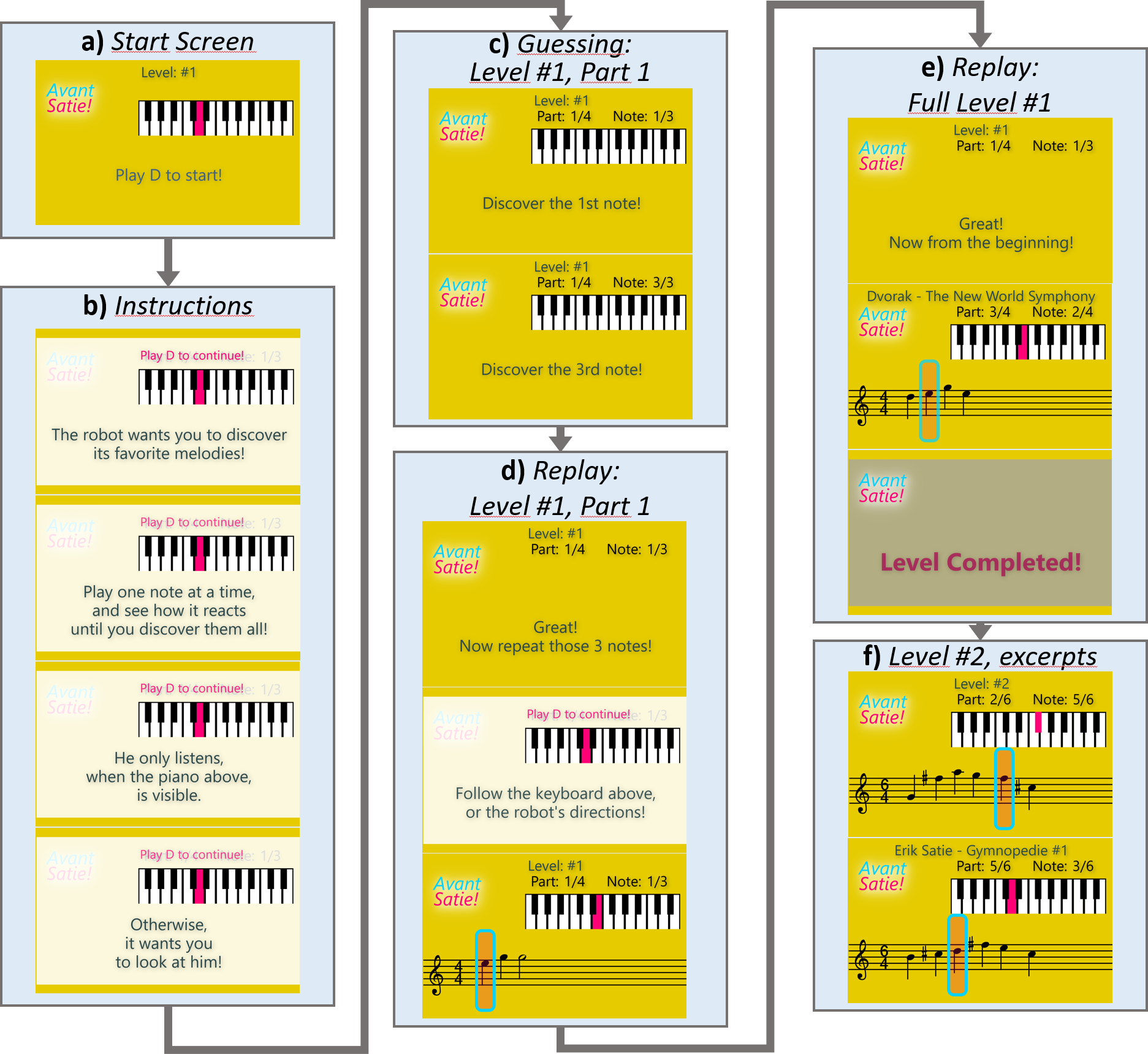}
	\caption{Various screenshots of the projected screen of the AvantSatie game: \textbf{a)} start screen; \textbf{b)} (part of) instruction screens; \textbf{c)} guessing phase in Level \#1; \textbf{d)} replay phase level \#1 (including replay phase instruction); \textbf{e)} Full-level replay and completion; \textbf{f)} Excerpts of replay phases for level \#2.}
	\label{fig:avantSatieScreens}
\end{figure}

Figure \ref{fig:avantSatieScreens} shows some shots of the AvantSatie screen throughout the game.
Here we present shots of the English version, while the study was ran using the Portuguese version (as can be seen by comparing these with Figure \ref{fig:avantSatieStudy}).

The start screen requires the player to interpret a piano figure and to interact with the floor piano (fig. \ref{fig:avantSatieScreens}.a), upon which the robot performs an affirmative animation, i.e., nodding as if trying to say "yes!" (first positive feedback).
This ensures that the player understands the basic interaction pattern of the activity i.e., screen displays instructions, playing the piano triggers a reaction on the robot.
It then follows with a little storyline and instructions on how to play (fig. \ref{fig:avantSatieScreens}.b).
Because Adelino is designed and animated as a non-verbal character, we rely on the screen to present in-game instructions, which also helps to immerse the player (versus having provided instructions prior to the activity).
Whenever an instruction screen is being presented (e.g. fig. \ref{fig:avantSatieScreens}.b) the robot turns to face the screen, 
as a mechanism to direct the player's attention to it (otherwise due to enthusiasm, the player might be too focused on the robot and overall set-up).
This also adds a feeling of presence - the robot is aware both of the player and of its surroundings (i.e. the screen - a point of shared attention).
When the first instruction set is over, the robot turns back towards the user and plays the affirmative animation again.
This animation is later used throughout the game, so it was important to initially present and reinforce it as a positive feedback.

On the first stage of all, the player is presented with no information except for the instruction "Discover the $1^{st}$ note!" (fig. \ref{fig:avantSatieScreens}.c).
While the player is moving around in from of the piano, the robot only performs face-tracking behaviour.
Upon playing some note, the robot assesses it as the player's guess. 
If it is correct, the robot performs the affirmative animation, after which the screen progress updates to e.g. "Discover the $2^{nd}$ note!", and the robot goes back to face-tracking. 
By having the robot provide feedback before the screen does, we manage to keep the player's visual attention focused on the message that the robot is communicating, instead of instigating them to search for feedback and new information on the screen (which we found to be the \textit{instinct} of most people, during our pilot tests and iterative scenario design process).

These steps repeat until all the notes of the current Stage are found.
After that,
the robot replays all the Stage's notes, while pointing at each corresponding piano key,
and then instructs the user to repeat it, with the screen exhibiting an illustration of the piano, highlighting each note, so that the player can unequivocally follow (fig. \ref{fig:avantSatieScreens}.d).
This Replay phase serves as a reward to the player for having struggled to discover the composition.
Each level was purposely built as a sequence of \textit{Challenging} followed by \textit{Rewarding} phases in order to maintain the user's engagement. 
At the end of each level the player gets to replay the full Level's composition as a bigger reward.
The name of the piece and composer is revealed, and the player replays all parts (fig. \ref{fig:avantSatieScreens}.e-f).
The first Level is slightly shorter and easier than the second one (no black keys are used).It is also used as a tutorial, and allows to separate data collection in order to mitigate the effects of learning how to interact with and play the game.

\subsection{Experimental Design}
The main purpose of this experiment was to study if ERIK successfully endowed Adelino with the ability to communicate task-directed information fully through its posture, while it is gaze-tracking the player.
For that, we compared three versions of AvantSatie.

In version \textbf{C-ERIK} and \textbf{C-EBPS}, Adelino would respond to each of the player's guesses by modifying its posture based on a "warm-cold" heuristic.
Upon each wrong guess the robot would therefore shift its expressive posture to either \textit{Warm} or \textit{Cold}, while keeping the gaze-tracking behaviour towards the player.
Figure \ref{fig:avantSatiePostures} (teaser in the first page) shows how these three postures look like on Adelino.
The figure shows each of the neutral, hot and cold posture as output by the ERIK algorithm while gazing forwards, towards an angle of about 25$^{\circ}$ above the horizon.


We did not initially mention that the robot would perform this type of behaviour to any participant, in any study condition.
Instead, it was expected that during the first minutes of the game (i.e. the Tutorial level), the players in these two conditions would notice that the robot was performing some behaviour that seemed congruent with their guesses, and would learn how to interpret the robot's posture in order to score better in the game.

The difference between \textbf{C-ERIK} and \textbf{C-EBPS} is purely technological - in \textbf{C-ERIK} we used ERIK to perform the gaze-tracking with an expressive posture.
EBPS, is a non-IK example-based posture synthesis technique for which we previously authored a large number of postures for each \textit{Warm}, \textit{Neutral} and \textit{Cold} expressions, each posture representing a pair (\textit{expression}, \textit{direction}), which sets the robot facing through a range of directions that are expected for this game, and produces a final posture through interpolation.
In run-time, given the face-tracking information, we take the two postures that represent the directions that are closest to the target one and interpolate them.
In our pilot studies we verified that this technique provided very smooth and acceptable results, by creating postures for each horizontal direction from $-70^{\circ}$ to $70^{\circ}$ with $10^{\circ}$ interval between them, which leads to 15 postures per expression, for each vertical direction. 
We initially considered to use 3 or 4 vertical directions, but upon testing realized that 2 would be enough, and that the interpolation between the two extreme vertical positions yielded acceptable results. 
The two vertical directions for which the postures were created were at $0^{\circ}$ (looking straight ahead, i.e., to the horizon) and $80^{\circ}$ (upwards).
The postures used in \textbf{C-EBPS} were additionally designed with special attention in order to look very similar to the ones generated by ERIK.

We note that, while the end-result looked similar to the one we wanted to achieve with ERIK, it required a considerable amount of work to create all those postures; 
Furthermore if we wanted to modify one of the expressions, we would be required to re-create all the postures for that expression and rebuild the interpolation model. 
Using ERIK only requires to author one single posture per intended expression.

Finally, \textbf{C-Control} differs from the other two conditions by not performing any posture-expressive behaviour at all.
The whole game is exactly the same, and the robot also performs gaze-tracking behaviour, using ERIK to perform gaze-tracking, but always holds the same, \textit{Neutral} posture throughout the game.
Therefore in this condition players rely solely on trial-and-error to discover the notes.

The study followed a between subjects design with random assignment within the three different conditions.
In all conditions the game design follows the same structure, with the same levels and tutorial information.


In order to address our research question, we established the following hypotheses:
\begin{itemize}
	\item \textbf{H1}: Players in \textbf{C-Control} will commit more mistakes than in \textbf{C-ERIK} and in \textbf{C-EBPS}.
	\item \textbf{H2}: Players in \textbf{C-ERIK} will play as well as in \textbf{C-EBPS}.
	\item \textbf{H3}: The robot is perceived to convey the illusion of life in all conditions.
	\item \textbf{H4}: The robot is perceived to convey the illusion of life more in \textbf{C-ERIK} and \textbf{C-EBPS} than in \textbf{C-Control}.
	\item \textbf{H5}: The players are able to perceive the robot's intention and motivation as being towards helping them in both \textbf{C-ERIK} and \textbf{C-EBPS} but not in \textbf{C-Control}.
	\item \textbf{H6}: The game is understandable and the robot is perceived to understand and play well in all conditions.
\end{itemize}

Upon arrival, the participants filled out the consent form in a separate room before being led to the game room. 
There they were given the same initial instructions, without revealing that the robot would indicate the result of their guesses through a change in posture. 
Instead they were solely informed that there would be two simple compositions to discover, 
that they should perform each guess and observe the robot, until they were able to discover all the notes, 
and otherwise just follow the instructions on screen.
The researcher would direct them to enter the room and start interacting without following them, as the robot was already active and would start face-tracking them once they stepped into the Kinect's field of view.
This ensured that the participants also noticed it immediately as an autonomous entity and would become immersed into the game.
The screen provided the starting instruction, which was to play a D note on the floor piano, along with an icon showing explicitly which key that was.
Therefore it was the participant who took the step to initiate the game, while also ensuring that they understood the piano to be a controller for it.
When finished, the participants were taken back to the initial room, administered a set of questionnaires, and received a thank-you gift (a movie ticket) at the end.

\label{sec:avantsatie_measures}

The following questionnaires were used as subjective measures:
\begin{itemize}
\item \textit{Networked Minds} \cite{HarmsBiocca2004} scale (\textbf{NM}), from which we took the \textit{Perceived Message Understanding} (\textbf{PMU}) and the \textit{Co-Presence} (\textbf{CP}) dimensions;
\item \textit{Inclusion of Other in Self} \cite{Aron1992} scale to measure the closeness that the participants felt with the robot (measure \textbf{IOS});
\item \textbf{RoSAS} \cite{carpinellaEtal2017} scale to measure the perception of the robot's social attributes Competence (\textbf{RC}), Warmth (\textbf{RW} and Discomfort (\textbf{RD});
\item \textit{Perceived Adaptability} (\textbf{PA}) from the \textit{Almere model};
\item \textit{Robot's Performance \& Usability} (\textbf{RPU}) scale to measure how well the participants felt the robot to be able to play the game, and how well the overall gameplay was designed;
\item \textit{Robot's Intention \& Motivation} (\textbf{RIM}) scale to measure how much participants felt the robot was there to provide tips and how much it wanted to succeed in helping them;
\item \textit{Animation Illusion of Life} (\textbf{AIL}) scale to measure the illusion of life of the robot.
\end{itemize}

The questionnaires for the RPU, RIM and AIL scales were specifically designed to address our research question.
The RPU scale in particular is composed of two dimensions:\\
\indent\textit{Task Performance} (\textbf{TP}) measures how well the participants perceived the robot to know the game and perform the task well; \\
\indent\textit{Task Usability} (\textbf{TU}) measures how easy and intuitive the participants felt it was to understand the task and the game-play interaction with the robot and the screen.

The RIM scale is also composed of two dimensions: \\
\indent\textit{Robot's Intention} (\textbf{RI}) measures how much the participants felt that the robot was providing hints to them throughout the task;\\
\indent\textit{Robot's Motivation} (\textbf{RM}) measures how much the participants felt that the robot's intrinsic motivation (i.e. \textit{purpose}) was to help them (by providing hints).

Table \ref{tab:specific_questionnaire} lists the questions used for each of these measures. 

All the questionnaires were answered in a 6-point Likert scale except for the RoSAS, which was answered in a 9-point Likert scale, and the IOS measure which was answered in a 7-point pictorial scale.
The data presented in the results has already been corrected by inverting the scores in negative scales.
All items were shuffled to mask for their dimensions.

{\scriptsize{\renewcommand{\arraystretch}{1.3}
\ctable[botcap,caption = RPU\, RIM and AIL questionnaires.,label=tab:specific_questionnaire,pos=htbp,width=\columnwidth]
{r >{\setlength{\baselineskip}{0.7\baselineskip}}X}{}{
	\multicolumn{2}{c}{\small RPU (TP + TU) - Robot's Performance \& Task Usability}\\
	\cline{2-2}
    TP1.& The robot knew where each note of each music was.\\
    TP2.&The robot always understood what note I had played.\\
    TP3.&The robot knew each music very well.\\
    TP4.&The robot knew every music by heart.\\
	TU1.&I had to look at the screen to know what happened at each moment.\\
	TU2.&I wouldn't understand the game without looking at the screen.\\
	TU3.&The game screen had all the info I needed to understand the game.\\
	TU4.&I had to follow the screen to know what to do.\\
%
	\multicolumn{2}{c}{\small RIM (RI + RM) - Robot's Intention \& Motivation}\\
	\cline{2-2}
	RI1.&I wouldn't have understood the game without the robot.\\
	RI2.&I managed to find the correct notes thanks to the robot.\\
	RI3.&I wouldn't have discovered the musics without the robot's help.\\
	RI4.&The tips that the robot gave me helped me to find the correct notes.\\
	RI5.&The robot's tips were consistent with my attempts to find each correct note.\\
	RI6.&The robot gave me tips while I was trying to find each correct note.\\
	RI7.&I was able to understand if I was close or far from the correct note thanks to the robot's tips.\\
	RM1.&The robot wanted me to find the correct notes.\\
	RM2.&The robot wanted me to discover all of the musics.\\
	RM3.&The robot was enthusiastic with my attempts to find the correct notes.\\
	RM4.&The robot thought about helping me.\\
%
	\multicolumn{2}{c}{\small AIL - Animation Illusion of Life}\\
	\cline{2-2}
	AIL1.&The robot's movement was smooth and natural.\\
	AIL2.&The robot seemed to be alive.\\
	AIL3.&The robot reminded me of characters I know from movies.\\
	AIL4.&The robot's motion followed my rhythm.\\
	&
}}}

The following objective data was also collected during each session, and measured only during the \textit{Guessing} phases of the game:
\begin{itemize}
\item \textbf{Time} spent guessing;
\item \textbf{WrongHot} number of incorrect guesses which were however assessed as Hot;
\item \textbf{WrongCold} number of incorrect guesses which were however assessed as Cold;
\item \textbf{WrongTotal} aggregates WrongHot and WrongCold.
\end{itemize}

\subsection{Experimental Results}
A total of 59 university students (30 females and 29 males) 
with ages ranging from 18 to 35 (M = 22.78; SD = 3.96) were recruited.
From these, two were excluded due to not complying with the instructions,
thus yielding a total of 57 valid participants, which resulted in a balanced distribution of 19 participants per condition.
19\% of the participants had already interacted with a robot before once, and 37\% more than once.
42\% reported a low to no level of expertise playing some musical instrument, while 39\% reported an intermediate expertise, and 19\% an advanced expertise. As for experience reading sheet music, 63\% reported a low score, 25\% an intermediate score, and 12\% declared to be experts. 


Throughout this analysis, we will be considering an additional \textbf{C-Enhanced} group which averages the measures of both \textbf{C-ERIK} and \textbf{C-EBPS} in order to treat them both as a single group.


We started by using the Shapiro-Wilk test of normality to verify for which measures the data was normally distributed ($\rho$>0.05).
Although we have four groups, in our analysis we will compare the means of only two groups at a time, thus we used the Student's t-Test when the data is normally distributed in both groups being tested, and the Mann-Whitney U test otherwise.

\begin{figure}
	\centering
	\begin{subfigure}{\columnwidth}
		\centering
		\includegraphics[width=0.6\columnwidth]{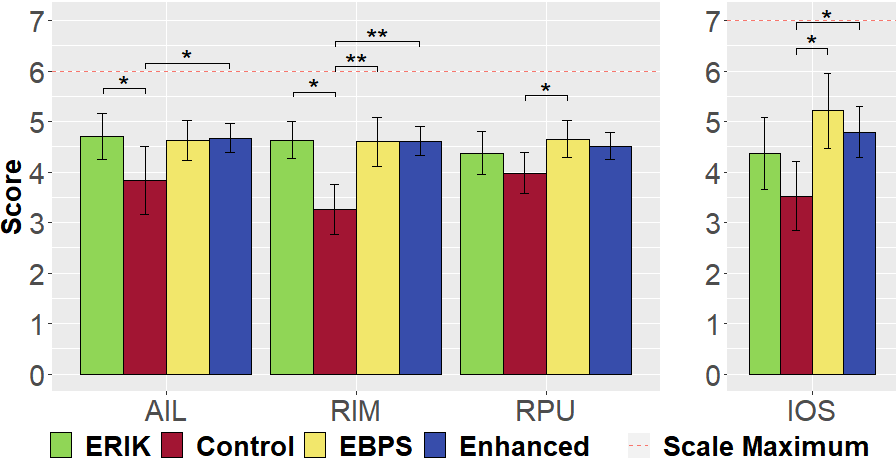}
		\caption{}~\label{fig:subjective_measures_scales}
	\end{subfigure}
	
	\begin{subfigure}{\columnwidth}
    	\centering
    	\includegraphics[width=0.6\columnwidth]{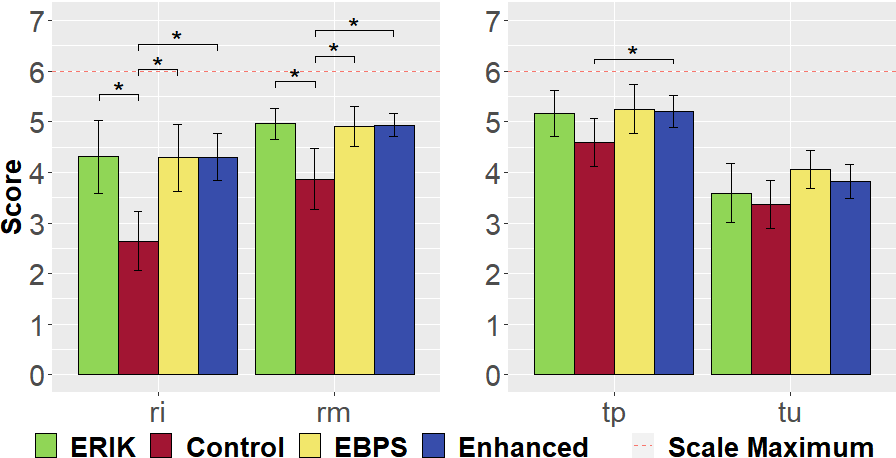}
    	\caption{}~\label{fig:subjective_measures_dimensions}
	\end{subfigure}
	
	\begin{subfigure}{\columnwidth}
	    \centering
	    \includegraphics[width=0.6\columnwidth]{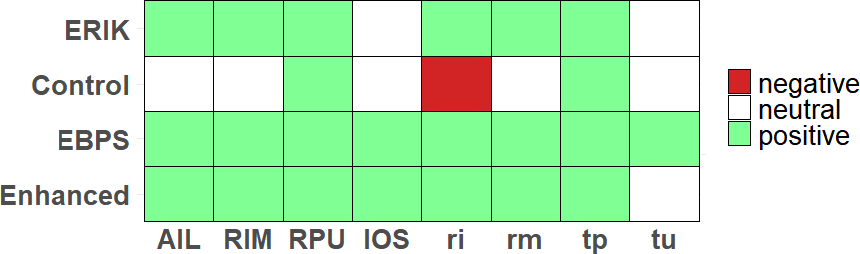}
	    \caption{}~\label{fig:subjective_measures_medians}
    \end{subfigure}
    
    \begin{subfigure}{\columnwidth}
    	\centering
    	\includegraphics[width=0.6\columnwidth]{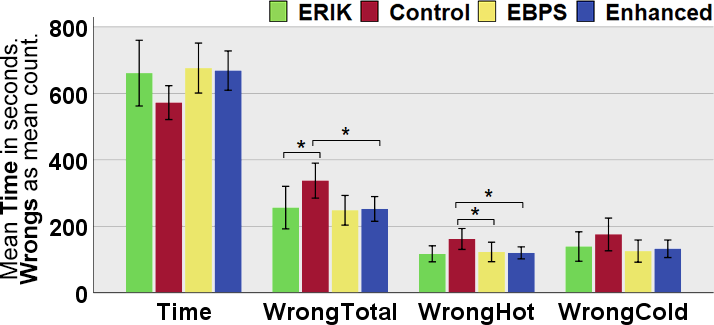}
    	\caption{}~\label{fig:objective_measures_results}
    \end{subfigure}
	\caption{\textbf{a)} Comparison of the scores of the subjective measures' scales. \textbf{b) Comparison of the dimensions used to compose some of the subjective measures. \textbf{c)} Comparison of means to the scale's median value on the subjective measures. \textbf{d)} Comparison of the objective measures. All error bars represent a 95\% C.I.}}~\label{fig:avantsatie_subjective_results}
\end{figure}

Table \ref{tab:subjective_comparison} shows the results of comparison of the means of all the subjective and objective measures, including sub-dimensions, between the different groups, with Figure \ref{fig:avantsatie_subjective_results} providing a clearer illustration.
\begin{table*}[ht]
\centering
\begin{tabular}{rrllll}
  \hline
 Measure Type & & ERIK-Control & EBPS-Control & Enhanced-Control & ERIK-EBPS \\ 
  \hline
  Subjective & AIL & \cellcolor[rgb]{0.50,1.00,0.58} ($\text{Tukey}, \rho=0.046*$) &  ($\text{Tukey}, \rho>0.05$) & \cellcolor[rgb]{0.50,1.00,0.58} ($\text{Tukey}, \rho=0.022*$) &  ($\text{Tukey}, \rho>0.05$) \\ 
  Subjective & RIM & \cellcolor[rgb]{0.50,1.00,0.58} ($\text{Dunn}, \rho=0.003*$) & \cellcolor[rgb]{0.50,1.00,0.58} ($\text{Dunn}, \rho=0.001*$) & \cellcolor[rgb]{0.50,1.00,0.58} ($\text{Dunn}, \rho=0.000*$) &  ($\text{Dunn}, \rho>0.05$) \\ 
  Subjective & RPU &  ($\text{TukeyHSD}, \rho>0.05$) & \cellcolor[rgb]{0.50,1.00,0.58} ($\text{Dunn}, \rho=0.030*$) &  ($\text{Dunn}, \rho>0.05$) &  ($\text{Dunn}, \rho>0.05$) \\ 
  Subjective & IOS &  ($\text{Tukey}, \rho>0.05$) & \cellcolor[rgb]{0.50,1.00,0.58} ($\text{Dunn}, \rho=0.009*$) & \cellcolor[rgb]{0.50,1.00,0.58} ($\text{Dunn}, \rho=0.044*$) &  ($\text{Dunn}, \rho>0.05$) \\ 
  \hline
Subjective (sub) & ri & \cellcolor[rgb]{0.50,1.00,0.58} ($\text{Dunn}, \rho=0.006*$) & \cellcolor[rgb]{0.50,1.00,0.58} ($\text{Tukey}, \rho=0.002*$) & \cellcolor[rgb]{0.50,1.00,0.58} ($\text{Dunn}, \rho=0.001*$) &  ($\text{Dunn}, \rho>0.05$) \\ 
Subjective (sub) & rm & \cellcolor[rgb]{0.50,1.00,0.58} ($\text{Welch}, \rho=0.002*$) & \cellcolor[rgb]{0.50,1.00,0.58} ($\text{Welch}, \rho=0.005*$) & \cellcolor[rgb]{0.50,1.00,0.58} ($\text{Welch}, \rho=0.002*$) &  ($\text{Welch}, \rho>0.05$) \\ 
Subjective (sub) & tp &  ($\text{Dunn}, \rho>0.05$) &  ($\text{Dunn}, \rho>0.05$) & \cellcolor[rgb]{0.50,1.00,0.58} ($\text{Dunn}, \rho=0.044*$) &  ($\text{Dunn}, \rho>0.05$) \\ 
Subjective (sub) & tu &  ($\text{Dunn}, \rho>0.05$) &  ($\text{Dunn}, \rho>0.05$) &  ($\text{Dunn}, \rho>0.05$) &  ($\text{Tukey}, \rho>0.05$) \\ 
  \hline
  Objective & Time & ($\text{Tukey}, \rho>0.05$) & ($\text{Tukey}, \rho>0.05$) &  ($\text{Dunn}, \rho>0.05$) & ($\text{Tukey}, \rho>0.05$) \\ 
  Objective & WrongTotal & \cellcolor[rgb]{0.50,1.00,0.58} ($\text{Dunn}, \rho=0.037*$) & ($\text{Dunn}, \rho>0.05$) & \cellcolor[rgb]{0.50,1.00,0.58} ($\text{Dunn}, \rho=0.015*$) & ($\text{Dunn}, \rho>0.05$) \\ 
  Objective & WrongHot & ($\text{Dunn}, \rho>0.05$) & \cellcolor[rgb]{0.50,1.00,0.58} ($\text{Dunn}, \rho=0.037*$) & \cellcolor[rgb]{0.50,1.00,0.58} ($\text{Dunn}, \rho=0.012*$) & ($\text{Dunn}, \rho>0.05$) \\ 
   \hline
\end{tabular}
\caption{Results of the subjective scales and sub-dimensions, and objective measures, including the post-hoc used in each case. Green with an asterisk* marks comparisons which are significantly different. Omitted measures (RW, RD and WrongCold) do not show any significant difference.} 
\label{tab:subjective_comparison}
\end{table*}
Within it we present each measure's comparison of means (Figures \ref{fig:subjective_measures_scales},\ref{fig:subjective_measures_dimensions},\ref{fig:objective_measures_results}) along with Figure \ref{fig:subjective_measures_medians} which shows how each subjective measure in each group compares to the scale's median value, i.e., is the average score significantly \textit{positive} (above median), \textit{negative} (below median), or \textit{neutral} (inconclusive).
For that we took each scale's median value (e.g., for the 6-point likert scale 'AIL' the median is $\frac{6+1}{2}=3.5$), and compared the distribution of each group's results to it.
We used the One-Sample t-Test when the data followed a normal distribution, and the One-Sample Wilcoxon Signed Rank test otherwise.
In order to make the presentation of results more summarized and easier to follow, we have also gathered the following comparison cases:
\begin{description}
\item[Strong Expressivity Difference (\textit{Strong E-D})]: Significant difference between \textbf{C-Control} and all \textbf{C-ERIK}, \textbf{C-EBPS} and \\\textbf{C-Enhanced} groups, with no difference between \textbf{C-ERIK} and \textbf{C-EBPS}. 
These findings will be attributed to the inclusion of the postural/intention-expressive behaviour in the activity, regardless of its technical implementation.
\item[Soft Expressivity Difference (\textit{Soft E-D})]: Significant difference between \textbf{C-Control} and \textbf{C-Enhanced}, and also between \\\textbf{C-Control} and either \textbf{C-ERIK} or \textbf{C-EBPS}.
These findings will also be attributed to the postural/intention-expressive behaviour, but with an indication that one of the technological implementations may have performed better.
\item[Technological Difference (\textit{T-D})]: Significant difference between the \textbf{C-Control} and either the \textbf{C-ERIK} or the \textbf{C-EBPS}, but not between the \textbf{C-Control} and the aggregated \textbf{C-Enhanced} group.
These findings will be attributed to some difference in the technological implementation only.
\end{description}

\subsubsection{Regarding the Subjective Measures}

Within the various subjective measures used in our study, we draw the following:
\begin{description}
	\item[AIL, RIM, NM and PA] all reported significantly lower scores in \textbf{C-Control} with a \textbf{\textit{Strong E-D}}. 
	\item[RPU] reported a significantly lower score in \textbf{C-Control} with a \textbf{\textit{Soft E-D}}.
	\item[IOS, RC] reported a significantly lower score in \textbf{C-Control} compared to \textbf{C-EBPS} with a \textbf{\textit{T-D}}.
	\item[RC] reported a significantly lower score in \textbf{C-Control} compared to \textbf{C-ERIK} with a \textbf{\textit{T-D}}.
	\item[RPU, NM, RC and RD] all reported positive scores in all groups.
	\item[AIL and RIM] reported positive scores in all enhanced groups, and neutral in the \textbf{C-Control} group.
	\item[PA, IOS and RW] are inconclusive regarding the polarity of the scores, given that they show mixed results.
\end{description}

Within the dimensions that compose these scales, we can verify as expected based on the \textbf{RIM} and \textbf{NM} results, that the \textbf{ri}, \textbf{rm}, \textbf{cp} and \textbf{pmu} dimensions all report the same \textbf{\textit{Strong E-D}}.
However the \textbf{RPU} scale had reported only a \textbf{\textit{Soft E-D}}.
Analysing further, we can see that the \textbf{tp} dimension (robot's Task Performance) did exhibit the expected \textbf{\textit{Strong E-D}}, while the \textbf{tu} dimension (Task Usability) reported only a significant \textbf{\textit{T-D}} between the \textbf{C-Control} and the \textbf{C-EBPS} conditions.

We also have a very interesting finding within the \textbf{RIM} scale. Regarding the Robot's Intention (\textbf{ri}), the \textbf{C-Control} scored significantly negative while all the other groups scored positively (and not neutral).
This is a very strong difference (which had already been pointed out in the comparison of means).
Similarly, albeit with a smaller difference, the Robot's motivation (\textbf{rm}) and the Networked Minds' Perceived Message Understanding (\textbf{pmu}) were also perceived to be neutral in \textbf{C-Control}, while in all the others it was positive.
Finally it is important to note that while the \textbf{RPU} scale was positive across all groups, we found that this was mostly due to the perceived robot's Task Performance (\textbf{tp}), as the Task Usability (\textbf{tu}) dimension scored significantly neutral on all groups, except on \textbf{C-EBPS} where it scored positively.

\subsubsection{Regarding the Objective Measures}
Figure \ref{fig:objective_measures_results} shows how the four objective measures performed across the different groups.
The measure of \textbf{WrongCold} did not reveal any differences between conditions.
However both the measures of \textbf{WrongHot} and \textbf{WrongTotal} show a \textbf{\textit{Strong E-D}} with the \textbf{C-Control} condition containing a significantly higher amount of wrong answers than the other groups.
The Time measure shows a \textbf{\textit{Soft E-D}} as the \\\textbf{C-Control} group performed significantly faster than the \textbf{C-EBPS} group ($t$=2.406, $\rho$=0.022) and the aggregated \textbf{C-Enhanced} group ($U$=242.0, $\rho$=0.044).



\subsection{Discussion and Conclusion}
The results collected and analysed show us in general that our research question is supported.
Looking into each of our initial hypotheses:

\indent\textbf{H1} - Players in \textbf{C-Control} will play worse than in \textbf{C-ERIK} and in \textbf{C-EBPS}: \textbf{True}. Participants in both \textbf{C-ERIK} and \textbf{C-EBPS} committed less mistakes than in the \textbf{C-Control},
by decoding and exploiting the hints given by the robot through its full-body posture.

\indent\textbf{H2} - Players in \textbf{C-ERIK} will play at least as well as in \textbf{C-EBPS}: \textbf{True}. As there was no significant difference between the total amount of wrong answers given by the players in \textbf{C-ERIK} and in the \textbf{C-EBPS}, thus the expressive postures provided by ERIK were as relevant and legible as the manually designed ones.

\indent\textbf{H3} - The robot is perceived to convey the illusion of life in all conditions: \textbf{Partially True}. All conditions except for \textbf{C-Control} reported a positive \textbf{AIL} average score. \textbf{C-Control} was neutral however, and not negative.
Our interpretation is that the robot's additional intentional-directed expressive behaviour in the enhanced groups contributed significantly to convey they illusion of thought (which is core to the illusion of life).

\indent\textbf{H4} - The robot is perceived to convey the illusion of life more in \textbf{C-ERIK} and \textbf{C-EBPS} than in \textbf{C-Control}: \textbf{True}. \noindent Both \textbf{C-ERIK} and \textbf{C-EBPS} scored significantly higher in the \exhyphenpenalty=10000 measure than the \textbf{C-Control} group.

\indent\textbf{H5} - The players are able to perceive the robot's intention and motivation as being towards helping them in both \textbf{C-ERIK} and \textbf{C-EBPS} but not in \textbf{C-Control}: \textbf{True}. The \exhyphenpenalty=10000 score was significantly higher and positive both in \textbf{C-ERIK} and \textbf{C-EBPS} than in \textbf{C-Control}.
When analysing RIM's sub-dimensions separately (\textbf{ri} and \textbf{rm}), we find that the Robot's Intention scored negative in the \textbf{C-Control} condition. 
This was very relevant as it was the only dimension that scored negatively.
Regarding the Robot's Motivation, we find a similar pattern, except that in \textbf{C-Control} it scored neutral (slightly better).
Our guess is that in general the players had a positive feeling about the robot (based on the results from RoSAS) and therefore, maybe considered that the robot did intrinsically want to help them (although they reported negatively on its hint-providing intention).
Furthermore, the Networked Minds (\textbf{NM}) and Almere's Perceived Adaptability (\textbf{PA}) scales both follow the same pattern, significantly higher in the enhanced conditions compared to \textbf{C-Control}, thus reinforcing that regardless of the technique used, the intention-directed behaviour of the robot, as designed and integrated into the gameplay, had a positive effect on various measures regarding the perception of the robot's intention, motivation and closeness towards the player.

\indent\textbf{H6} - The game is understandable and the robot is perceived to play well in all condition: \textbf{Partially True}. The \textbf{RPU} measure was positive in all groups. 
However when breaking down the scale, we find the Task's Usability (\textbf{tu}) was scored as neutral in all except the \textbf{C-EBPS} group, and that this difference is actually significant compared to \textbf{C-Control}.
The game was designed in order to ensure that the participants would not become too affected by the lack of the posture-expressive behaviour (in \textbf{C-Control}) that they would not understand the task at all.
While the \textbf{tu} measure reports a \textbf{T-D} on \textbf{C-EBPS}, it was not reported in the whole \textbf{C-Enhanced}, which means that we fail to refute that the inclusion or absence of the robot's intention-directed expressive behaviour does not cause a significant effect on the participants' understanding of the game and the task.
Therefore the iterative game-design method (with 3 pilot tests) allowed to tweak the usability of the game to an acceptable level, even in the absence of the robot's full expressive behaviour, while also noting that the overall game-design and/or interaction design could have still been made better.
\\

We further noted additional findings such as that participants in \textbf{C-Enhanced} took more time to complete the task than in \textbf{C-Control}.
Inspection of the audio-video data captured from the study revealed that participants in the former conditions, 
having understood that the robot was giving tips, would try notes at a lower pace in order to inspect the robot's response, 
while in \textbf{C-Control}, after a while they would quickly play random notes, driving quicker through the task with a significantly higher number of mistakes.

The RoSAS scale shows only a significant difference in the Robot Competence (\textbf{RC}) measure between \textbf{C-ERIK} and \textbf{C-Control}.
However the difference did not hold for the whole of the \textbf{C-Enhanced} group.
We suspect that the ERIK algorithm may have yielded a higher feeling of competence, because the use of that algorithm is prone to result in more dynamic/responsive motion, which players may have attributed to a higher sense of acknowledgement of the other, and capability of attention, on the robot's part.
In overall however, no concrete difference may be concluded between \textbf{C-ERIK} and \textbf{C-EBPS}, given that on comparing the various scales, there were either none, or mixed differences (e.g. in contrast to the previous remarks, for the IOS scale, the \textbf{C-EBPS} scores significantly higher than \textbf{C-Control}, but here \textbf{C-ERIK} does not).
Although the RoSAS scale seems not to have added any relevant information, that fact may be used to also hypothesize that across all groups, the robot was perceived as being nearly the same \textit{character} - which was desirable for our study.


We have found evidence that our research question (Section \ref{sec:avantsatie}) is positively supported, given that:
\begin{itemize}
	\item When ERIK was used, the participants noted its intention\-directed postural behaviour, and were able to intuitively understand it without having been given any information about its existence in order to perform better on a task that required it.
	\item The effect of using ERIK was similar to that of a manually tailored (and laborious) alternative technique EBPS, in that no significant differences were found for any of the measures between those two groups, while significant ones were found especially on key measures when compared to the Control condition. This means that ERIK can be used in substitution of such currently existing manually-tailored and arduously worked techniques (such as ones based on \textit{learning-by-examples}).
	\item The difference between the various conditions did not hinder the player's understanding and playability of the game, impacting only on their performance, which reveals that the selected task was properly designed to answer our question.
	\item Despite the shakiness of the robot due to the fact of it being a low-fidelity craft robot, results show that the algorithm succeeded in making it convey both the intended expressivity, and the illusion of life, meaning that it is likely to work on both similar or more solid robots.
\end{itemize}

In addition, we highlighted the importance of the \textit{illusion of thought} to the overall \textit{illusion of life in robots}, as already had been initially proposed by Takayama et al. \cite{Takayama2011}.
In our case we further demonstrated that such illusion can also be expressed through fully interactive expressive postures that are computed in real-time, and are therefore most appropriate for use-cases involving autonomous social robots.

\section*{ACKNOWLEDGEMENTS}
This work was supported by national funds through FCT - Funda\c{c}\~{a}o para a Ci\^{e}ncia e a Tecnologia with references UID/CEC/50021/2019 and SFRH/BD/97150/2013.
\bibliographystyle{unsrt}  
\bibliography{bib/software,bib/PhD,bib/tese_robots_manual_entries,bib/inpress,bib/arxivNutty/main,bib/arxivERIK/main}
\clearpage

\end{document}